\pdfoutput=1
\documentclass[11pt]{article}

\usepackage[]{acl}
\usepackage{float}
\usepackage{times}
\usepackage{latexsym}
\usepackage{times}
\usepackage{adjustbox}
\usepackage{latexsym}
\usepackage{algorithm}
\usepackage{algpseudocode}
\usepackage{todonotes}

\usepackage{multirow}
\usepackage{subcaption}
\usepackage{booktabs}
\usepackage[flushleft]{threeparttable}

\usepackage{hyperref}
\hypersetup{colorlinks,allcolors=darkblue}

\usepackage{graphicx}
\usepackage{xcolor}
\usepackage{tabularx,booktabs}
\usepackage{float}
\usepackage[T1]{fontenc}
\usepackage[utf8]{inputenc}

\usepackage{microtype}

\title{On Robust Incremental Learning over Many Multilingual Steps}

\author{Karan Praharaj \and Irina Matveeva \\
        Reveal \\ Chicago, IL\\
        \texttt{\{kpraharaj,imatveeva\}@revealdata.com}}
        
\begin{document}
\maketitle
\begin{abstract}
Recent work in incremental learning has introduced diverse approaches to tackle catastrophic forgetting from data augmentation to optimized training regimes. However, most of them focus on very few training steps. We propose a method for robust incremental learning over dozens of fine-tuning steps using data from a variety of languages. We show that a combination of data-augmentation and an optimized training regime allows us to continue improving the model even for as many as fifty training steps. Crucially, our augmentation strategy does not require retaining access to previous training data and is suitable in scenarios with privacy constraints. 
\end{abstract}
\section{Introduction}

Incremental learning is a common scenario for practical applications of deep language models. In such applications, training data is expected to arrive in batches rather than all at once, and so incremental perturbations to the model are preferred over retraining the model from scratch every time new training data becomes available for efficiency of time and computational resources. When multilingual models are deployed in applications, they are expected to deliver good performance over data across multiple languages and domains. This is why it is desirable that the model keeps acquiring new knowledge from incoming training data in different languages, while preserving its ability on languages that were trained in the past. The model should ideally keep improving over time, or at the very least not deteriorate its performance on certain languages through the incremental learning lifecycle.

It is known that incremental fine-tuning with data in different languages leads to catastrophic forgetting \cite{french-catastrophic, mccloskey:catastrophic} of languages that were fine-tuned in the past \cite{liu_preserving_2021, vu-overcoming-catastrophic}. This means that the performance on previously fine-tuned tasks or languages decreases after training on a new task or language. Multiple strategies have been proposed to mitigate catastrophic forgetting. Data-focused strategies such as augmentation and episodic memories \cite{hayes-aug-strategy-example-1, arslan-tiny-episodic, lopez-paz_gradient_2017}, entail maintaining a cache of a subset of examples from previous training data, which are mixed with new examples from the current training data. The network is subsequently fine-tuned over this mixture as a whole, in order to help the model "refresh" its "memory" of prior information so that it can leverage previous experience to transfer knowledge to future tasks.

\begin{figure}[t!]
  \centering
\includegraphics[width=\columnwidth, scale=1.5]{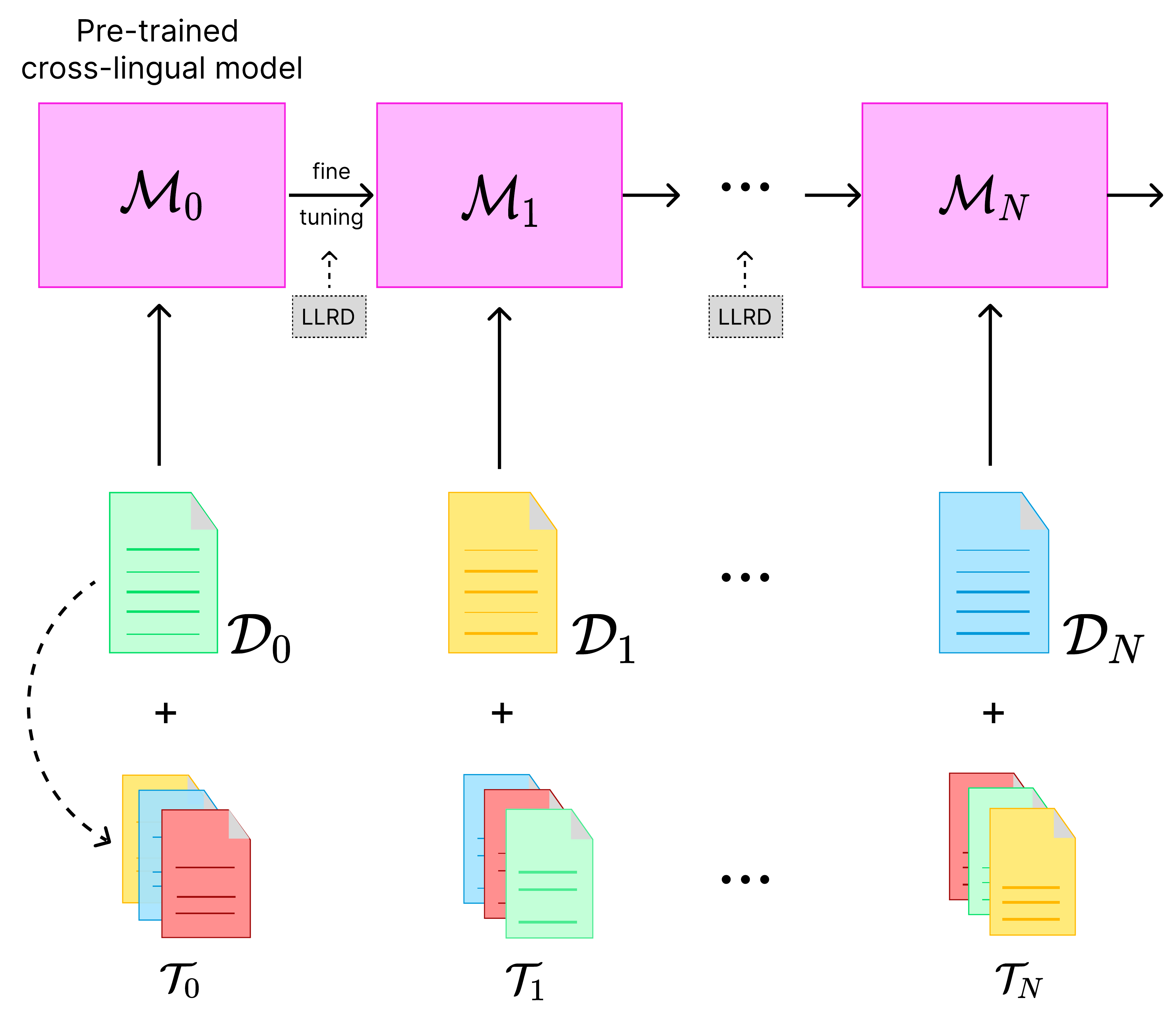}
  \caption{Translation augmented sequential fine-tuning approach with LLRD-enabled training. We begin with a pre-trained multilingual model $\mathcal{M}_{0}$ and fine-tune it over multiple stages to obtain ($\mathcal{M}_{i}$ where $i$ = 0...$N$). At each fine-tuning stage, we train the model $\mathcal{M}_{i}$ over examples from the training set  that is available at that stage ($\mathcal{D}_{i}$) which is in language $\mathcal{L}_{j}$. At each step, we sample a small random subset from $\mathcal{D}_{i}$ and translate that sample into languages ($\mathcal{L}$ $\backslash$ $\mathcal{L}_{j}$) to create a set of translated examples $\mathcal{T}_{i}$. Each step of training includes LLRD as a hyperparameter.}
  \label{fig:1}
\end{figure}

Closely related to our current work is the work by  \citet{mhamdi_cross-lingual_2022, ozler-2020} of understanding the effect of incrementally fine-tuning models with multi-lingual data. They suggest that joint fine-tuning is the best way to mitigate the tendency of cross-lingual language models to erase previously acquired knowledge. In other words, their results show that joint fine-tuning should be used instead of incremental fine-tuning, if possible.

Optimization focused strategies such as \citet{training-regime-mirzadeh-2020, kirkpatrick_overcoming_2017} focus on the training regime, and show that techniques such as dropout, large learning rates with decay and shrinking the batch size can create training regimes that result in more stable models.

Translation augmentation has been shown to be an effective technique for improving performance as well. \citet{neubig-translation, Fadaee_2017-trans-aug, liu-etal-trans-aug} and \citet{xia-etal-2019-trans-aug} use various types of translation augmentation strategies and show substantial improvements in performance. Encouraged by these gains, we incorporate translation as our data augmentation strategy.

In our analysis, we consider an additional constraint that affects our choice of data augmentation strategies. This constraint is that the data that has already been used for training cannot be accessed again in a future time step. We know that privacy is an important consideration for continuously deployed models in corporate applications and similar scenarios and privacy protocols often limit access of each tranche of additional fine-tuning data only to the current training time step. Under such constraints, joint fine-tuning or maintaining a cache like \citet{chaudhry_tiny_2019, lopez-paz_gradient_2017} is infeasible. Thus, we use translation augmentation as a way to improve cross-lingual generalization over a large number of fine-tuning steps without storing previous data. 

In this paper we present a novel translation-augmented sequential fine-tuning approach that mixes in translated data at each step of sequential fine-tuning and makes use of a special training regime. Our approach shows minimization of the effects of catastrophic forgetting, and the interference between languages. The results show that for incremental learning over dozens of training steps, the baseline approaches result in catastrophic forgetting. We see that it may take multiple steps to reach this point, but the performance eventually collapses.

The main contribution of our work is combining data augmentation with adjustments in training regime and evaluating this approach over a sequence of 50 incremental fine-tuning steps. The training regime makes sure that incremental fine-tuning of models using translation augmentation is robust without the access to previous data. We show that our model delivers a good performance as it surpasses the baseline across multiple evaluation metrics. To the best of our knowledge, this is the first work to provide a multi-stage cross-lingual analysis of incremental learning over a large number of fine-tuning steps with recurrence of languages.

\section{Related Work}

Current work fits into the area of incremental learning in cross-lingual settings. \citet{mhamdi_cross-lingual_2022} is the closest work to our research. The authors compare several cross-lingual incremental learning methods and provide evaluation measures for model quality after each sequential fine-tuning step. They show that combining the data from all languages and fine-tuning the model jointly is more beneficial than sequential fine-tuning on each language individually. We use some of their evaluation protocols but we have different constraints: we do not keep the data from previous sequential fine-tuning steps and we do not control the sequence of languages. In addition, they considered only six hops of incremental fine-tuning whereas we are interested in dozens of steps. \citet{ozler-2020} do not perform a cross-lingual analysis, but study a scenario closely related to our work. Their findings fall in line with those of \citet{mhamdi_cross-lingual_2022} as they show that combining data from different domains into one training set for fine-tuning performs better than fine-tuning each domain separately. However, this type of joint fine-tuning is ruled out for our scenario where we assume that access to previous training data is not available, and so we focus on sequential fine-tuning exclusively.

\citet{training-regime-mirzadeh-2020} study the impact of various training regimes on forgetting mitigation. Their study focuses on learning rates, batch size, regularization method. This work, like ours, shows that applying a learning rate decay plays a significant role in reducing catastrophic forgetting. However, it is important to point out that our type of decay is different from theirs. \citet{training-regime-mirzadeh-2020} start with a high initial learning rate for the first task to obtain a wide and stable minima. Then, for each subsequent task, slightly decrease the learning rate, while simultaneously reducing the batch size, as recommended by \citet{dont-decay-lr-inc-batch-size}. On the other hand, we apply our decay rate across the transformer model's layer stack so that the deviations from the current optimum get progressively smaller as one moves down the layers and we do this at each step of incremental fine-tuning.

 Memory-based approaches such as \citet{arslan-tiny-episodic, lopez-paz_gradient_2017} have been explored to mitigate forgetting. Such methods make use of an \textit{episodic memory} or a cache which stores a subset of data from previous tasks. These examples are then used for training along with the current examples in the current optimization step. Similarly, \citet{xu-etal-2021-gradual} suggest a gradual fine-tuning approach, wherein models are eased towards the target domain by increasing the concentration of in-domain data at every fine-tuning stage. This work builds on the findings from \citet{bengio-curriculum}, who show that a multi-stage curriculum strategy of learning easier examples first, and gradually increasing the difficulty level leads to better generalization and faster convergence.  While we cannot maintain a cache of this sort because of our constraints, we take inspiration from this line of research and generate ``easier examples'' using translation in languages that are expected to appear in our data.

 Sequential fine-tuning of languages has not been extensively studied for long sequences. \citet{liu_preserving_2021} and \citet{garcia_towards_2021} go up to two stages, whereas \citet{mhamdi_cross-lingual_2022} go upto six stages. We provide an analysis of a much longer fine-tuning sequence with fifty stages. We are also the first to present an analysis of sequences with repetition of languages.

\section{Method}

We propose a translation augmented sequential fine-tuning approach for incremental learning in a cross-lingual setting. Our approach addresses the scenario in which a pre-trained model is incrementally fine-tuned over dozens of steps without access to previously seen training data. There is a set of languages $\mathcal{L} = \mathcal{L}_{0}$, ..., $\mathcal{L}_{K}$ that can appear during the incremental fine-tuning steps and we assume that in each step the data comes from only one language. We exploit the benefits of data augmentation, as well as specialized optimization techniques. 

We begin with a pre-trained multilingual model $\mathcal{M}_{0}$ which will be fine-tuned over multiple stages to create incremental versions $\mathcal{M}_{i}$ where $i$ = 0...$N$. The training data in each incremental fine-tuning step is $\mathcal{D}_{i}$ and is in a randomly selected language $\mathcal{L}_{j}$, where $ 0 \leq j \leq K$. At each step, we sample a small random subset  $\mathcal{T}_{i}$ from $\mathcal{D}_{i}$ and translate that subset to all languages from $\mathcal{L}$ except $\mathcal{L}_{j}$, to create multiple additional subsets of training data $\mathcal{T}_{i}$. 
We denote the augmented training set as $\mathcal{D}_{i}^{\mathcal{T}}$, where $$\mathcal{D}_{i}^{\mathcal{T}} = \mathcal{D}_{i} + \mathcal{T}_{i}$$

Figure \ref{fig:1} provides a graphical representation of our approach.

\subsection{Fine-tuning regime with LLRD}
Motivated by \citet{yosinski_how_2014}, we apply a layer-wise learning rate decay (or LLRD, denoted by $\zeta$) to the model parameters depending on their position in the layer stack of the model, based on the discriminative fine-tuning method proposed by \citet{howard-ruder-2018}.
Layer-wise Learning Rate Decay (LLRD) is a method that applies higher learning rates for top layers and lower learning rates for bottom layers. The goal is to modify the lower layers that encode more general information less than the top layers that are more specific to the pre-training task. This is accomplished by setting the learning rate of the top layer and using a multiplicative decay rate to decrease the learning rate layer-by-layer from top to bottom. We split the parameters $\theta$ into $\left\{\theta^1, \cdots, \theta^L\right\}$ where $\theta^l$ contains the parameters of the $l^{th}$ layer of the pre-trained model. The parameters are updated as follows:
$$\theta_t^l=\theta_{t-1}^l-\eta^l \cdot \nabla_\theta J(\theta)$$
where $\eta^l$ represents the learning rate of the $l^{th}$ layer. We set the learning rate of the top layer to $\eta^l$ and use
$$\eta^{k-1}=\zeta \cdot \eta^k$$
\section{Experiments}

\subsection{Data}
We use the Multilingual Amazon Reviews corpus (MARC) \cite{keung-etal-2020-multilingual}. This dataset is a large-scale collection of product reviews from 6 different languages and from 31 different categories. We construct our training sets by extracting reviews for ten common categories: \textit{apparel, automotive, beauty, drugstore, grocery, home, kitchen, musical instruments, sports, wireless}.
The number of reviews for each language-category combination are not equal, but to ensure consistency of training examples at each training step, we create two unique training sets of size 100 and 150 for each language-category combination.
For our experiments, we use reviews from all 6 languages provided in the dataset (Chinese, English, French, German, Japanese and Spanish). We drop the 3-star reviews and bifurcate the rest into two class labels: positive (4-star and 5-star) and negative (1-star and 2-star) sentiment.\footnote{We ensure that both final labels contain an equal number of examples of their constituent star-ratings. E.g., the negative sentiment class will contain an equal number of reviews from 1- and 2-star reviews.} 

Each incremental training set $\mathcal{D}_{i}$ contains 100 reviews from a particular language-category combination, for example, \textit{de-grocery}. To ensure class balancing, we sample an equal number of positive and negative records for each training set.  

We use the original test-splits for each of the 60 language-category combinations of the MARC data as our test set. 

\subsection{Translation augmentation}
The translations were generated  using the Google Translate API. In the current work we sample a fraction of 0.1 of the training examples. For example, if the training data is $\mathcal{D}_{i}$ has 100 records, the translation augmented data $\mathcal{D}_{i}^{\mathcal{T}}$ will have 150 records. 

\subsection{Constructing the sequence}
We tested our approach on a large number of incremental fine-tuning steps using data from various language-category combinations. To do that we created 3 random sequences $\mathcal{S}_{1}$, $\mathcal{S}_{2}$, $\mathcal{S}_{3}$ with 50 training sets $\mathcal{D}_{1}$-$\mathcal{D}_{50}$ each. We term each incremental fine-tuning step as a \textit{hop}.  Multiple hops comprise a \textit{sequence}.

Each training set $\mathcal{D}_{i}$ contains data from a particular language-category combination. To construct these 3 randomized sequences we used the following approach. We first generated all possible language-category combinations, and then sampled one combination at a time for each hop. The only constraint placed on the sampling is that it cannot choose a combination that has already appeared in the sequence. However, the same language with a different category and the same category with a different language still can occur. E.g., if \textit{de-grocery} features once in the sequence, it cannot be repeated, even though \textit{de-sports} or \textit{en-grocery} are possible options. The plots in Fig. \ref{fig:2} and \ref{fig:3} show the language-category combination for each $\mathcal{D}_{i}$ in all three sequences.

\subsection{Model and training}
We use multilingual BERT (cased) as our base model. We run a hyperparameter optimization over a relatively small search space containing values that were most effective in our preliminary experiments. 
Two types of settings were used for training:
\begin{itemize}
    \item Default settings: 
        \begin{itemize}
            \item Epochs: 5
            \item Learning rate: $2e-5$
            \item LLRD: 1.0
        \end{itemize}
        \item LLRD-enabled settings: 
        \begin{itemize}
            \item Epochs: 5
            \item Learning rate: $2e-5$
            \item LLRD: 0.75, 0.85 \footnote{Our preliminary experiments showed that LLRD values of $\zeta=0.75$ and $\zeta=0.85$  are the most suitable candidates for our scenario. We show in brief the results of other values for comparison in Table \ref{tab:1}. The models are trained over the \textit{sports} datasets of all six languages and compare the scores averaged over the 60 evaluation sets. $\zeta=0.75$ and $\zeta=0.85$ deliver the most consistent performance.} 
        \end{itemize}
\end{itemize}

The best checkpoint from any given stage is chosen for subsequent fine-tuning over the next language dataset. At the first stage, we use the pre-trained mBERT checkpoints released by \cite{devlin-bert}.\footnote{\url{github.com/google-research/bert/blob/master/multilingual.md}}
All experiments have been run on a single machine with a 6-core NVIDIA Tesla K80 GPU.

\begin{table}[ht!]
  \centering
  \scalebox{0.97}{
  \resizebox{\linewidth}{!}{%
\begin{tabular}{ccccccccccccccccccc}
\toprule[1.5pt]
{\textbf{Train data} $\downarrow$}  &   {{$\mathbf{\zeta = 0.38}$}}  &   {$\mathbf{\zeta = 0.5}$}  &   {$\mathbf{\zeta = 0.75}$}  &   {$\mathbf{\zeta = 0.85}$}  &   {$\mathbf{\zeta = 0.95}$} &   {$\mathbf{\zeta = 1.0}$} \\

 \midrule[1.0pt]
 \textbf{de-100}    &   0.70    &   0.73     &   \textbf{0.76}    &   0.75     &   0.75   &   0.69 \\
 \textbf{en-100}    &   0.65        &   0.64    &   \textbf{0.73}    &   0.69     &   0.65    &   0.66 \\
 \textbf{fr-100} &   0.71    &   0.72    &   \textbf{0.73}    &   \textbf{0.73}    &   0.72      &   0.69 \\
 \textbf{jp-100} &      0.62    &    0.63      &   0.65    &   \textbf{0.67}    &   0.63    &   0.33 \\
 
 \textbf{zh-100} &   0.46    &   0.57     &   \textbf{0.69}    &   0.65     &   0.59   &   0.57 \\
 \textbf{es-100} &   0.70    &   0.73     &   \textbf{0.76}    &   0.75     &   0.69   &   0.63 \\
 
 \bottomrule[1.5pt]
 
\end{tabular}
}
}
  \caption{
  Comparison of different LLRD settings ($\zeta$). We observe that $\zeta=0.75$ and $\zeta=0.85$ deliver the most consistent performance. % 
  }\label{tab:1}
  \vspace{-1mm}
\end{table}

\subsection{Experimental setup}
We show the results with the following four variations. We start with the default incremental fine-tuning approach and add modifications such as translation augmentation, LLRD and the combination of translation and LLRD. 
\begin{itemize}
\setlength\itemsep{0.5em}
\item \textit{Sequential fine-tuning (\textsc{SeqFT})}: Data is $\mathcal{D}_{i}$, default training settings.
\item \textit{Sequential fine-tuning with LLRD (\textsc{SeqFT-Llrd})}: Data is $\mathcal{D}_{i}$, Trained with LLRD-enabled settings.
\item \textit{Translation augmented sequential fine-tuning (\textsc{SeqFT-Trans})}: Data is $\mathcal{D}_{i}^{\mathcal{T}}$, default training settings.
\item \textit{Translation augmented sequential fine-tuning using LLRD (\textsc{SeqFT-Trans-Llrd})}: This is our approach. Data is $\mathcal{D}_{i}^{\mathcal{T}}$, Trained with LLRD-enabled settings.

\end{itemize}

\subsection{Evaluation Metrics}

We evaluate our proposed approach against the baseline models on overall $F_{1}$ scores over the following metrics:
\begin{itemize}
    \item \textbf{Average hop-wise $F_{1}$}: The $F_{1}$ scores over each of the 60 test sets are averaged for every single fine-tuning hop. 
    \item \textbf{Overall $F_{1}$}: The averages of hop-wise $F_{1}$ scores for all stages are averaged  to give the overall performance.
    \item \textbf{Forgetting (F)}: The average forgetting across languages at the end of sequential fine-tuning. This evaluation metric measures the decrease in performance on each of the languages between the peak $F_{1}$ score and the $F_{1}$ score after final training step of the sequence. We evaluate forgetting by language (F-lang) as well as by category (F-categ). 
    \item \textbf{In-language, in-domain performance (IL/ID)}: These are the average scores on all the test sets corresponding to the last fine-tuned language-category combination. For example, if the current stage of fine-tuning uses Chinese \textit{zh-grocery} data, then the in-language performance is the $F_{1}$ over the \textit{zh-grocery} test set.
    \item \textbf{Out-of-language, in-domain performance (OL/ID)}: These are the average scores on all the test sets corresponding to languages that were \textit{not} seen in the previous stage of fine-tuning but are of the same domain. For example, if the current stage of fine-tuning uses \textit{zh-grocery} data, the test sets used to calculate OL/ID performance are English (\textit{en-grocery}), French (\textit{fr-grocery}), German (\textit{de-grocery}), Japanese (\textit{jp-grocery}) and Spanish (\textit{es-grocery}).
    \item \textbf{In-language, out-of-domain performance (IL/OD)}: The average scores on test sets of the same language as training but corresponding to the domains that were \textit{not} used during training.  For example, if the current stage of fine-tuning uses \textit{zh-grocery} data, the performance on the Chinese test sets of all domains other than \textit{grocery} are averaged at each fine-tuning stage.
    \item \textbf{Out-of-language, out-of-domain performance (OL/OD)}: The average scores on the test sets corresponding to all language-category combinations except the one that was used during training.
\end{itemize}

\section{Results}
\label{sec:results}

\begin{table*}[ht!]
\footnotesize
  \centering
    {%
\begin{threeparttable}

\begin{tabular*}{\textwidth}{llcccccccc}
\toprule
          \textbf{Method} & \textbf{Size} & \textbf{Seq} & \textbf{Overall $F_{1}$} & \textbf{IL/ID} & \textbf{OL/OD} & \textbf{IL/OD} & \textbf{OL/ID} & \textbf{F-lang} & \textbf{F-categ}   \\
\midrule
\midrule
    
    \textbf{\textsc{SeqFT}} &    100 &     \multirow{4}{*}{1} &  35.36 &      35.96 &        35.35 &       35.11 &       35.34 &     39.32 &         39.32  \\
    
    \textbf{\textsc{SeqFT-Llrd}} &     100 &      &  85.44 &      85.70 &        85.44 &       85.94 &       85.24 &     2.73 &         2.66  \\
    
    % \textbf{\textsc{SeqFT-Llrd}} &     150 &      &  86.53 &      87.36 &        86.52 &       87.41 &       86.58 &     1.59 &         1.57 &        1.69 &            0.91 \\
    
    \textbf{\textsc{SeqFT-Trans}} &  100+50 &      &  43.54 &      43.18 &        43.55 &       43.91 &       42.74 &     40.50 &         39.21 \\
    
    \textbf{\textsc{SeqFT-Trans-Llrd}} &     100+50 &      &  83.05 &      84.86 &        83.02 &       84.31 &       82.93 &     1.54 &         1.53  \\
    
    \midrule
    
    \textbf{\textsc{SeqFT}} &     100 &     \multirow{4}{*}{2} &  40.71 &         40.86 &           40.71 &          42.39 &          40.46 &        38.70 &            35.83 \\
    
    \textbf{\textsc{SeqFT-Llrd}} &     100 &      &  85.32 &         86.39 &           85.30 &          86.03 &          85.09 &         1.37 &             1.33 \\
    
    % \textbf{\textsc{SeqFT-Llrd}} &     150 &      &  86.04 &         87.22 &           86.02 &          86.98 &          85.86 &         2.06 &             1.78 &            0.54 &                0.74 \\
    
    \textbf{\textsc{SeqFT-Trans}} &  100+50 &      &  39.94 &         39.21 &           39.95 &          40.55 &          39.74 &        43.78 &            43.37  \\
    
    \textbf{\textsc{SeqFT-Trans-Llrd}} &     100+50 &      &  84.67 &         86.45 &           84.64 &          86.00 &          84.62 &         2.04 &             1.75 \\

    \midrule
    
    \textbf{\textsc{SeqFT}} &     100 &     \multirow{4}{*}{3} &  48.74 &         49.73 &           48.72 &          49.58 &          48.65 &        48.37 &            48.10  \\
    
    \textbf{\textsc{SeqFT-Llrd}} &     100 &      &  84.49 &         86.04 &           84.46 &          85.44 &          83.99 &         1.31 &             1.24 \\
    
    % \textbf{\textsc{SeqFT-Llrd}} &     150 &      &  85.65 &         86.52 &           85.63 &          86.31 &          85.44 &         2.15 &             1.73 &            0.51 &                1.80 \\
    
    \textbf{\textsc{SeqFT-Trans}} &  100+50 &      &  35.51 &         35.18 &           35.52 &          36.15 &          35.39 &        37.04 &            36.02  \\
    
    \textbf{\textsc{SeqFT-Trans-Llrd}} &     100+50 &      &  84.74 &         86.02 &           84.71 &          85.74 &          84.42 &         0.65 &             0.82\\

\bottomrule
\end{tabular*}
\caption{
  Summary of results of our approach in comparison to the baselines. 
  }
\smallskip
\scriptsize
\label{tab:2}
\end{threeparttable}

}

\end{table*}%

We present below a comparison of our approach \textsc{SeqFT-Trans-Llrd} with different variations of sequential fine-tuning in our results. 

\begin{figure*}[htp!]\centering
\clearpage
\includegraphics[scale=0.3]{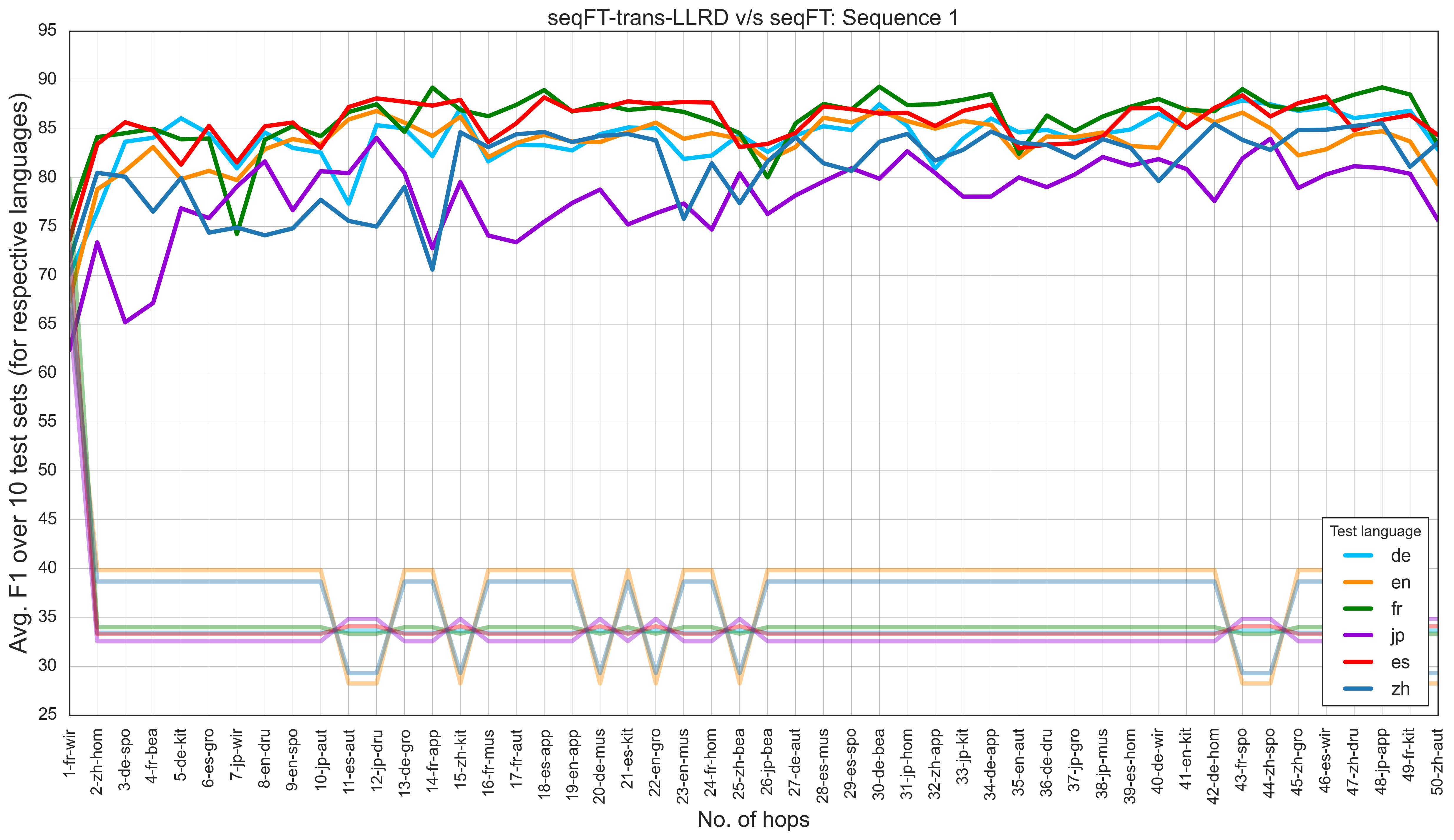}

\includegraphics[scale=0.3]{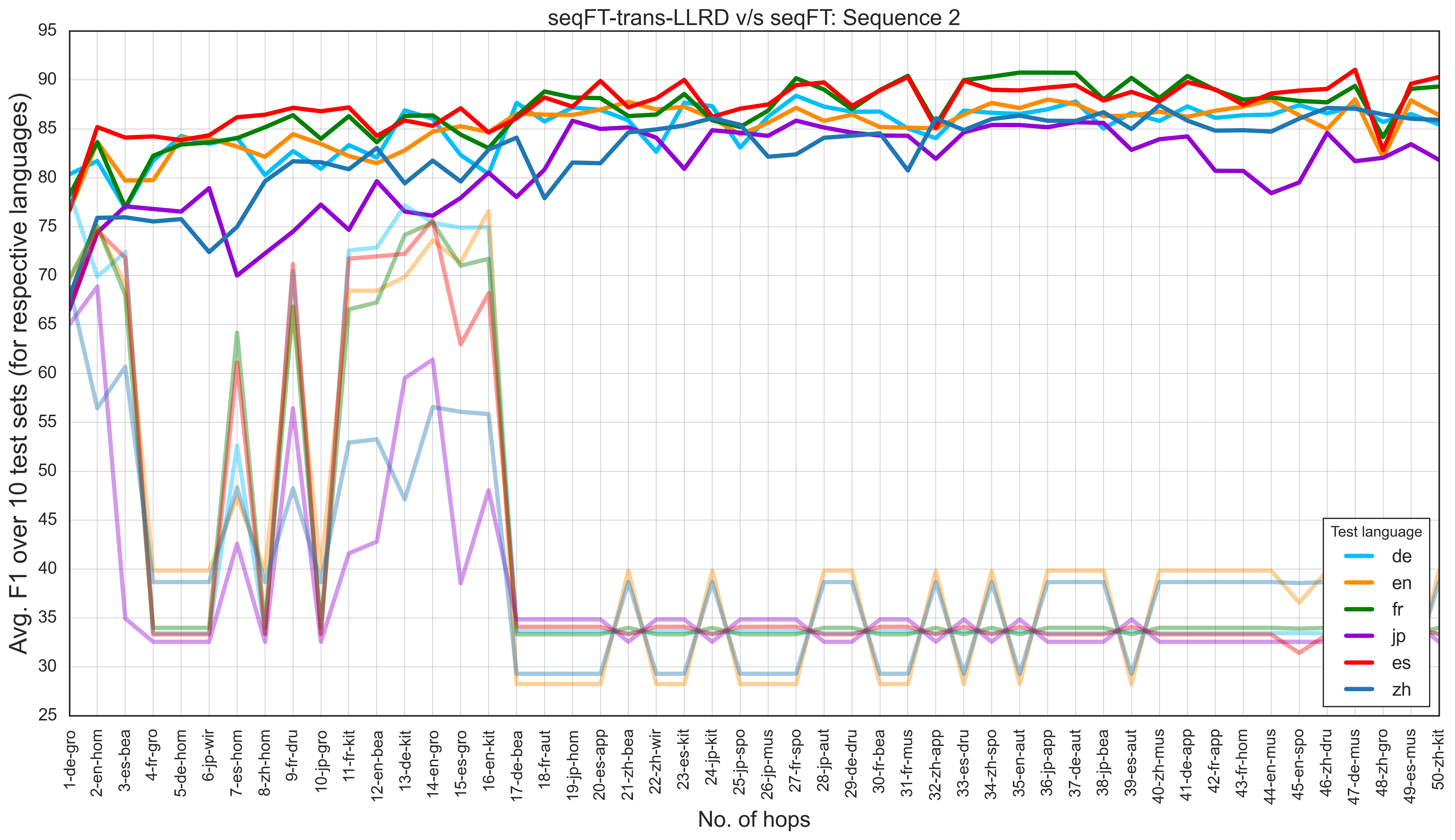}

\includegraphics[scale=0.3]{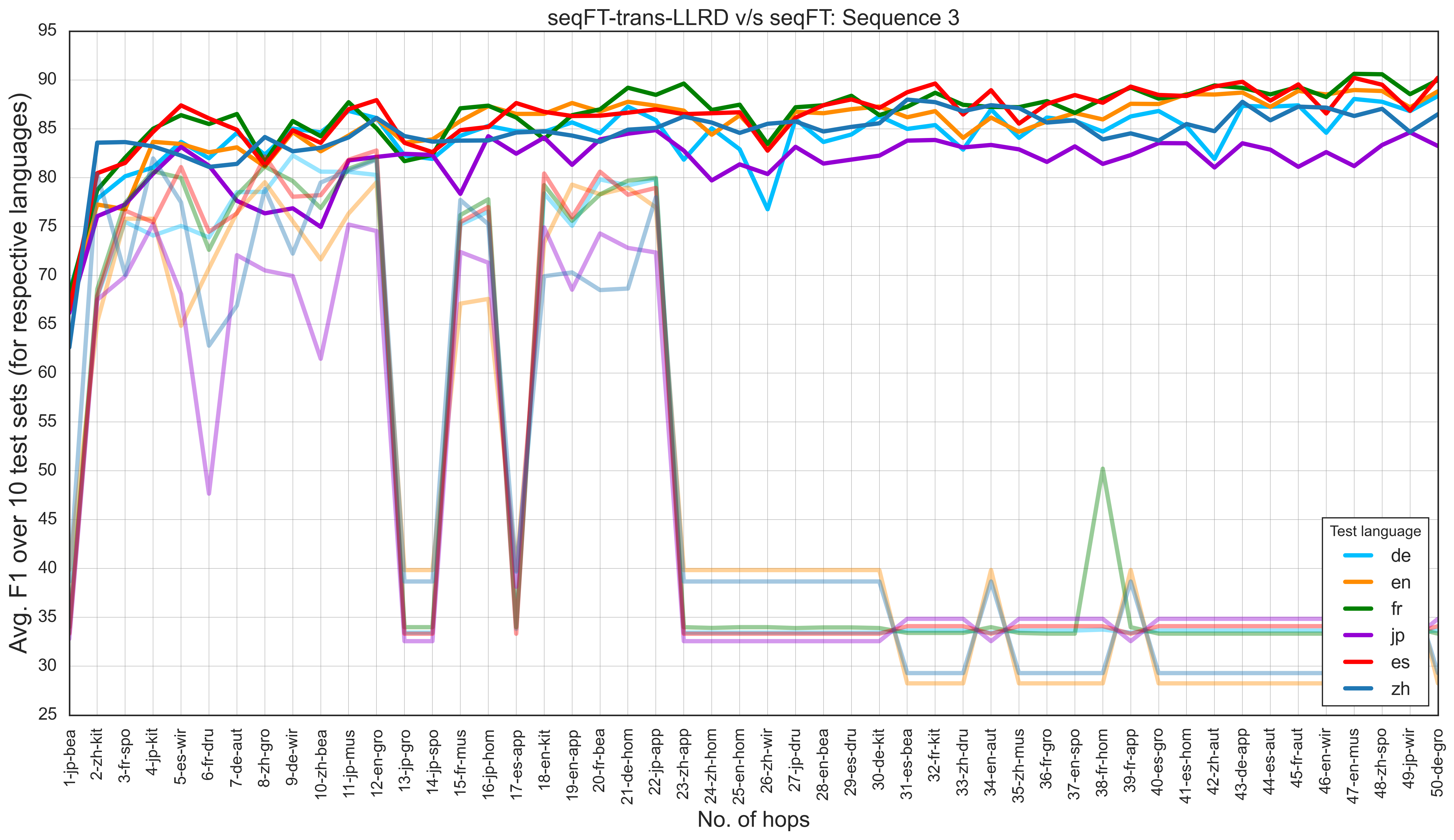}

\caption{ \textsc{SeqFT} vs \textsc{SeqFT-Trans-Llrd}: We show the plots of hop-wise $F_{1}$ scores for each randomized sequence of 50 hops each. Each plot has the details for one sequence. We show the $F_{1}$ for each language separately in color-coded lines. The translucent  lines show the results for our baseline of no data augmentation and default fine-tuning settings. The regular lines show the results of our approach. The x-axis shows the language-category combination in each training set $D_{i}$.}
\label{fig:2}
\end{figure*}

\begin{figure*}[htp!]\centering
\clearpage
\includegraphics[scale=0.3]{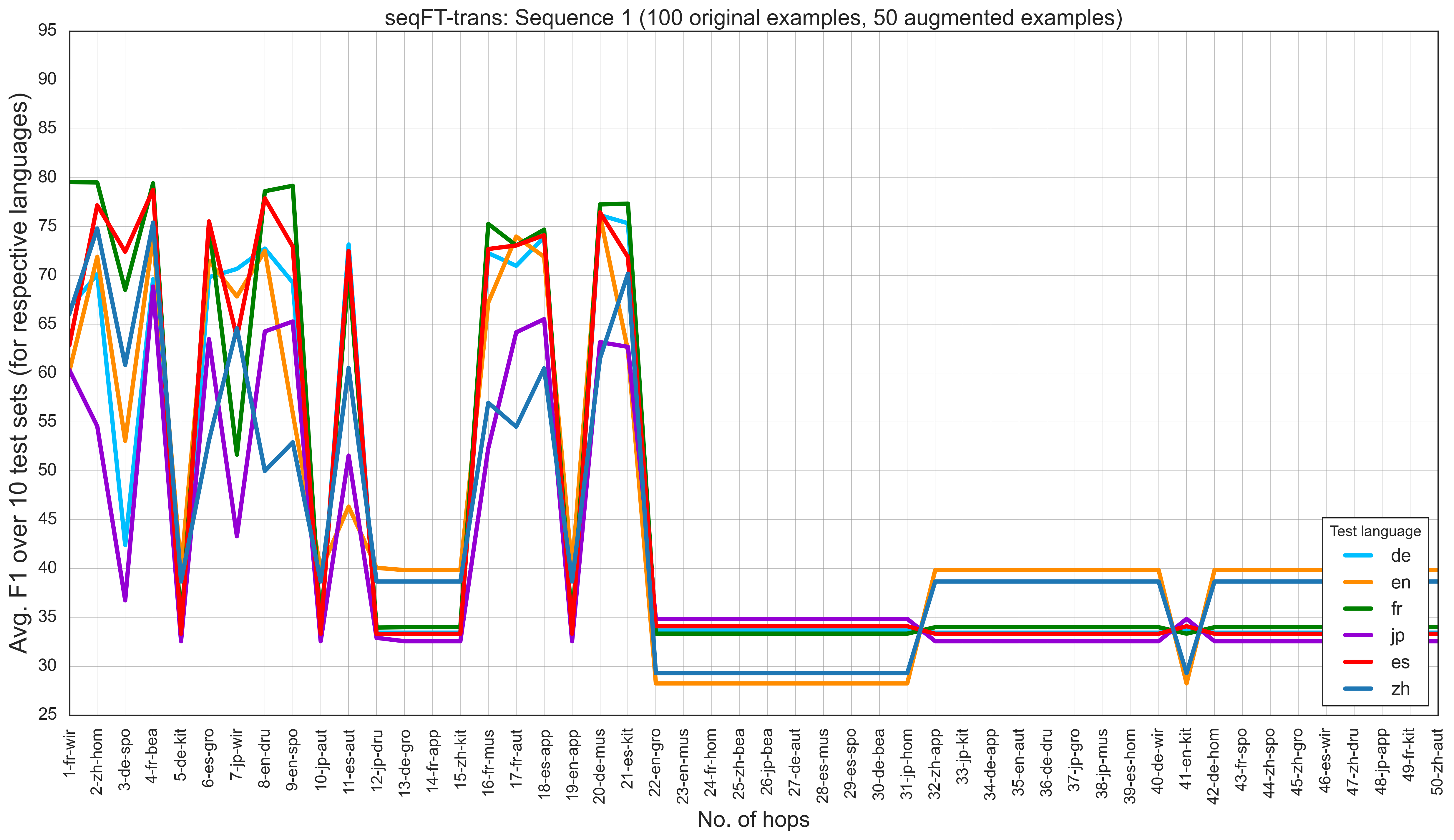}
\includegraphics[scale=0.3]{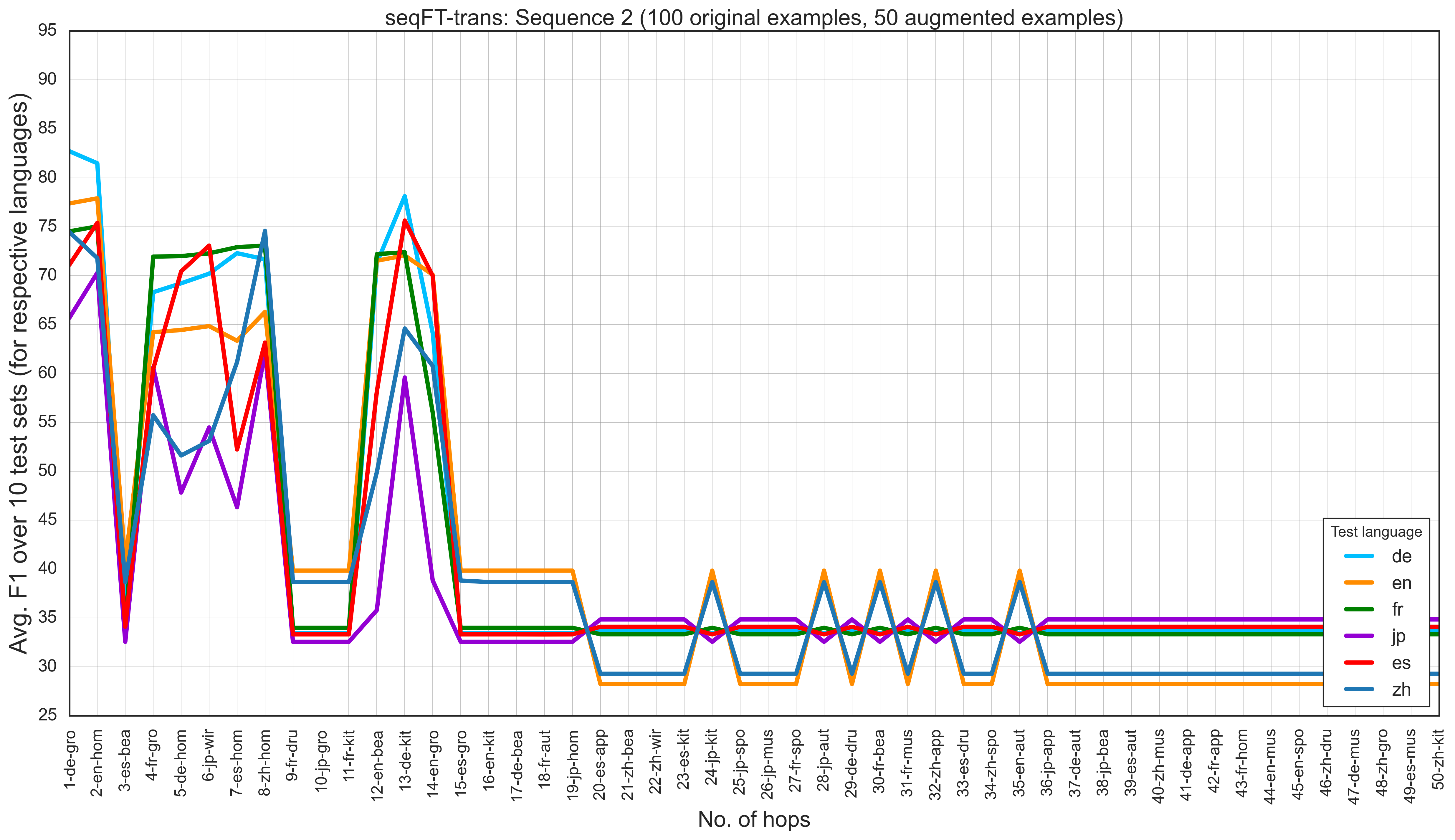}
\includegraphics[scale=0.3]{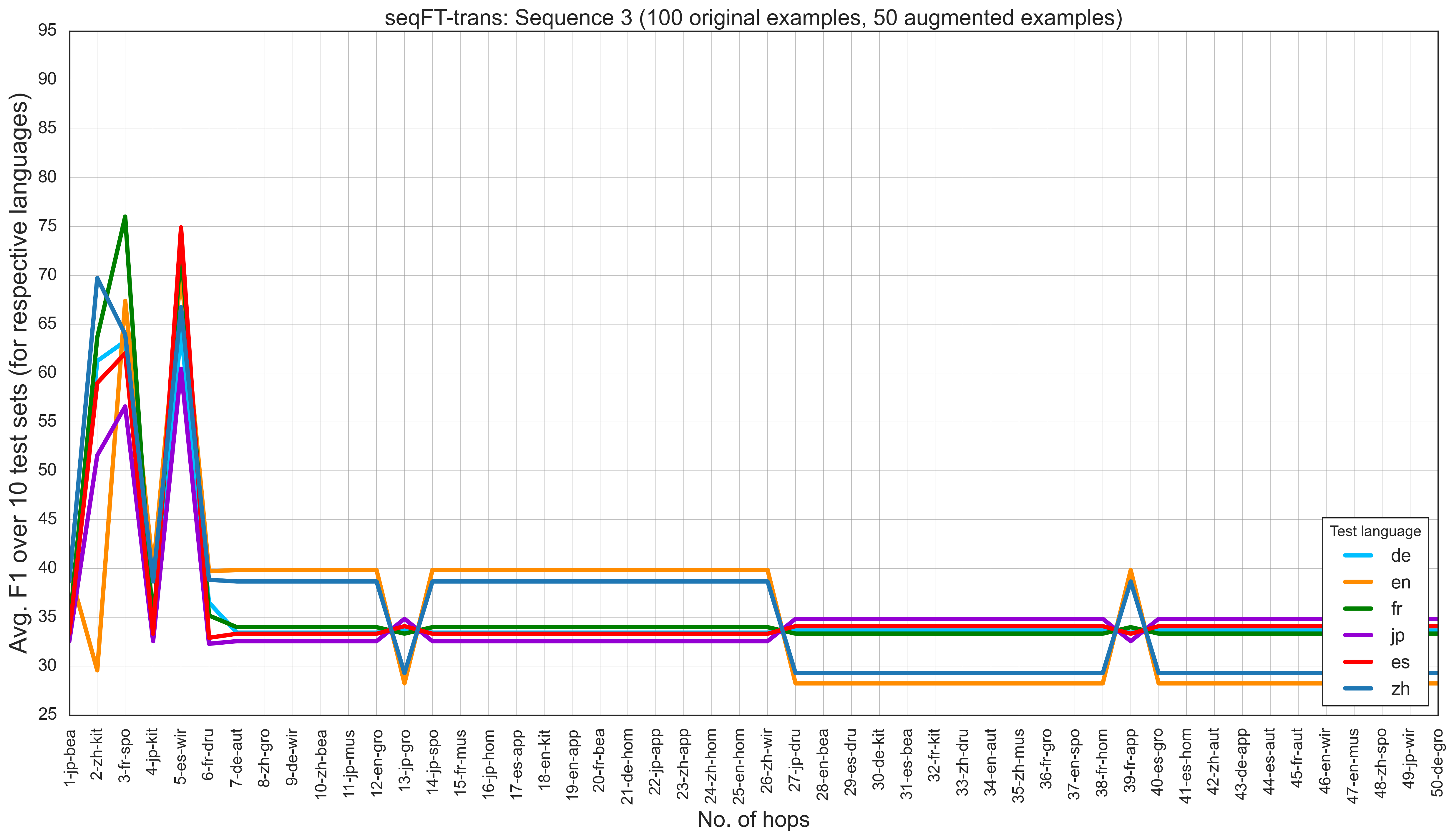}

\caption{ \textsc{SeqFT-Trans}: We show the plots of hop-wise $F_{1}$ scores for each randomized sequence of 50 hops each. Each plot has the details for one sequence. We show the $F_{1}$ for each language separately in color-coded lines. The x-axis shows the language-category combination in each training set $D_{i}$.}
\label{fig:3}
\end{figure*}

\subsection{ \textsc{SeqFT} \normalfont{(baseline)}} \label{subsec:seqFT}

Our proposed approach \textsc{SeqFT-Trans-Llrd} outperforms \textsc{SeqFT} decisively. We see that it is able to dramatically improve the overall $F_{1}$ performance and reduce forgetting on both forgetting metrics by one order of magnitude. This is evident from Fig. \ref{fig:2} and Table \ref{tab:2}. We see in the plots for average $F_{1}$ (Fig. \ref{fig:2}) that for each of the three sequences, the default approach \textsc{SeqFT} results in catastrophic forgetting. It can happen at different hops. In sequence 1, at the 2nd hop, in sequence 2, at the 17th hop and in sequence 3, at the 23rd hop. But eventually, the $F_{1}$ drops and never recovers. This highlights the importance of studying sequential fine-tuning over a large number of steps to be able to observe these effects. After the model performance collapses, we observed that the model classifies almost every example as negative. It is not clear from our results in these three different sequences if a particular language or category or combination triggers this collapse in performance. This is something we intend to explore in future work. Even in the initial hops before the collapse in performance, \textsc{SeqFT} under-performs our approach. From Table \ref{tab:2}, we see that our approach outperforms this baseline on all metrics. There is at least a 36 point improvement on the overall $F_{1}$ score between \textsc{SeqFT} and our approach.

In subsections \ref{subsec:seqFT-trans} and \ref{subsec:seqFT-LLRD}, we study the effects of the translation augmentation and specialized training regime separately to understand their contributions in isolation.  

\subsection{\textsc{SeqFT-Trans} \normalfont{(baseline)}} \label{subsec:seqFT-trans}

We observe that translation augmentation on its own performs very similarly to the baseline \textsc{SeqFT}. It outperforms the baseline only on one sequence in terms of overall $F_{1}$. The overall $F_{1}$ for \textsc{SeqFT-Trans} is significantly lower compared to our approach. The plots look similar to \textsc{SeqFT}, but we still provide them in Fig. \ref{fig:3}. Augmentation seems to delay catastrophic forgetting until 6, 15 and 22 hops. However, both approaches eventually result in catastrophic forgetting. Thus, the performance at the end of the sequence is extremely low. This is reflected in the high values of the F-lang and F-categ metrics in Table \ref{tab:2}.

\subsection{ \textsc{SeqFT-Llrd}} \label{subsec:seqFT-LLRD}

In contrast to \textsc{SeqFT-Trans}, using only the specialized training regime \textsc{SeqFT-Llrd} shows a strong performance. In fact, it appears that the main advantage of our approach stems from the optimized training regime with LLRD since \textsc{SeqFT-Llrd} and \textsc{SeqFT-Trans-Llrd} have comparable performance on many evaluation metrics. For sequence 3, our full approach \textsc{SeqFT-Trans-Llrd} shows a slightly higher overall $F_{1}$ performance as compared to \textsc{SeqFT-Llrd}. But on the other two sequences \textsc{SeqFT-Llrd} has a higher $F_{1}$ score. However, in terms of the forgetting metrics, our approaches outperforms \textsc{SeqFT-Llrd} on two out of three sequences. Also, for sequence 3, the out-of-domain performance with our full approach is higher.

In summary, our approach outperforms the baseline (\textsc{SeqFT}) by a wide margin. Since we use multiple language and category combinations, we show results on metrics based on similarity of the train and test data with respect to language and category. Our observations are consistent across all evaluation metrics. The main performance boost for our approach comes from including LLRD in the training regime. However, our combination of LLRD and translation augmentation slightly outperforms \textsc{SeqFT-Llrd} in terms of both forgetting metrics.

\section{Conclusion}
We introduce a sequential fine-tuning approach wherein the language data for fine-tuning is augmented by a subset of translated examples. Our augmentation strategy emulates episodic memory and decreases the reliance on a cache of stored examples from previous stages. We also advocate the use of layer-wise learning rate decay and illustrate its effectiveness in mitigating forgetting. With our results, we show that the proposed approach can outperform joint fine-tuning based methods, in spite of not having access to the complete set of examples from all languages. Crucially, it achieves robust and consistent performance over multiple cross-lingual fine-tuning stages. The trajectories of performances over different languages suggest that the model can continue learning over new data (or languages) for even more stages in the sequence without undergoing a significant reduction in performance. Furthermore, our approach surpasses all baselines when evaluated on in-domain, out-of-domain, in-language and out-of-language performance, showing that the model has a strong generalization ability. All-in-all, we hope our work provides encourament to the community to pursue similar recipes that facilitate long-term continual learning.

\section{Limitations}

One of the primary limitations of our work is that the analysis has only been provided for a single random sequence with only six languages. Diversification of this study with more languages, more random sequences and an even higher number of fine-tuning stages is a strong avenue for future work that we intend to pursue. Additionally, we would also like to extend this study to other cross-lingual tasks to see if the findings are similar. 

Another limitation is a lack of experimentation with adapter-based methods. In the future, we would also like to experiment with varying proportions of translated examples with respect to the original training size.

We would also like to extend our work to include a more in-depth study of the underlying linguistic factors that underpin cross-lingual transfer or forgetting. A study of this kind would ideally include, but not be limited to, analyses based on word order, scripts, morphology and syntax.

\section*{Acknowledgements}
The authors of this work would like to express their gratitude to Dinesh Karamchandani, for help with setting up the experimentation framework, and Dan Roth for feedback on our approach.

\bibliography{custom, main, anthology}
\bibliographystyle{acl_natbib}

\end{document}